\documentclass{article}

\usepackage{PRIMEarxiv}

\usepackage[utf8]{inputenc} 
\usepackage[T1]{fontenc}    
\usepackage{hyperref}       
\usepackage{url}            
\usepackage{booktabs}       
\usepackage{amsfonts}       
\usepackage{nicefrac}       
\usepackage{microtype}      
\usepackage{lipsum}
\usepackage{fancyhdr}       
\usepackage{graphicx} 
\usepackage{algorithm}
\usepackage{algorithmicx}
\usepackage{algpseudocode}
\usepackage{subcaption}
\graphicspath{{media/}}     

\title{\textbf{On the ability of CNNs to extract color-invariant intensity-based features for Image Classification}}

\author{Pradyumna Elavarthi$^{1}$,
        James Lee$^{2}$, Anca Ralescu$^{1}$ 
\thanks{$^1$ Department of Electrical Engineering and Computer Science, University of Cincinnati, OH 45221 USA.}%
\thanks{$^2$ Digital Scholarship Center, University of Cincinnati, OH 45221 USA.}%
}

\begin{document}
\maketitle

\begin{abstract}
Convolutional neural networks (CNNs) have demonstrated remarkable success in vision-related tasks. However, their susceptibility to failing when inputs deviate from the training distribution is well-documented. Recent studies suggest that CNNs exhibit a bias toward texture instead of object shape in image classification tasks, and that background information may affect predictions. This paper investigates the ability of CNNs to adapt to different color distributions in an image while maintaining context and background. The results of our experiments on modified MNIST and FashionMNIST data demonstrate that changes in color can substantially affect classification accuracy. The paper explores the effects of various regularization techniques on generalization error across datasets and proposes a minor architectural modification utilizing the dropout regularization in a novel way that enhances model reliance on color-invariant intensity-based features for improved classification accuracy. Overall, this work contributes to ongoing efforts to understand the limitations and challenges of CNNs in image classification tasks and offers potential solutions to enhance their performance. 
\end{abstract}


\section{Introduction}
Deep learning models have made significant advancements in various domains, particularly in vision and natural language processing \cite{imagenetclassificationwithcnns}. Among the popular architectures, convolutional neural networks (CNNs) have emerged as powerful tools for visual tasks, finding applications in diverse fields from medicine to construction. The success of CNNs can be attributed not only to the increased computational capacity but also to their remarkable generalization capabilities \cite{reconcilingmodern}. However, recent studies have uncovered potential challenges associated with the generalization of CNNs, revealing their susceptibility to learning non-causal features and their limited performance on distributions different from the training data \cite{invariantrisk}, \cite{limitationhosseini},\cite{shortcutlearning}, \cite{cognitive}.

The fundamental objective of machine learning is to develop models that not only fit the training data well but also demonstrate high performance on unseen test data, known as generalization. Overfitting occurs when a model becomes overly specialized to the training data, capturing spurious patterns that do not exist in the broader population, leading to poor generalization despite low training error. To address this issue, various regularization techniques have been devised to improve the model's generalization by mitigating the impact of overfitting. These techniques include weight penalization to discourage excessive importance on specific features and dropout regularization, which introduces randomness by probabilistically excluding neurons, thereby encouraging the model to rely on other informative features.\par 
While pooling operations in CNNs contribute to their translation invariance, data augmentation plays a crucial role in enhancing their overall robustness. However, studies have shown that certain distributional shifts, such as maintaining the shape while altering the texture of objects, can bias CNNs towards prioritizing texture over shape information.\cite{texturebiasgeirhos} proposed that improving the shape bias of models can enhance their robustness against such perturbations. Furthermore, investigations into the shape bias property of CNNs conducted by Hosseini et al. \cite{limitationhosseini} revealed that CNNs do not inherently exhibit a strong shape bias, leading to reduced accuracies in detecting negative images. The authors argued that negative images, which are semantically similar to normal images, can pose challenges for CNNs and demonstrated their limitations in this regard.\par 
In this research paper, we investigate the generalization ability of convolutional neural networks (CNNs) in classifying images sampled from different color distributions while preserving identical pixel intensities compared to the training data. Our study focuses on two representations of the dataset: a custom-colored version of MNIST \cite{mnistpaper} and a more complex dataset, FashionMNIST \cite{fashionmnist}, which is employed to validate the obtained results. Specifically, we modify the MNIST data to create three distinct datasets: one containing only green color in the first dataset, another with a single color channel in the second dataset, and a third dataset incorporating all three color channels. Notably, equal numbers of examples are assigned to each class label within each color variant, ensuring no inherent correlation between color and class label. As a result, these datasets are semantically similar, as color conveys no additional information for classification purposes.\par
To generate the datasets, we randomly assign a color channel (red, blue, or green) to each entire image in the custom MNIST dataset. Additionally, we divide the images into three horizontal parts and assign a random color to each part. This approach guarantees that all datasets contain an equal distribution of edges and shapes, enabling us to evaluate whether CNNs possess the capability to extract and utilize color-invariant features during the prediction process. Three identical Convnet architectures are initialized with the same weights and trained simultaneously on these datasets.\par
For testing purposes, a common test dataset is employed, consisting of images with all three color channels. This test dataset allows us to assess the performance of both models in handling images containing multiple color channels. By conducting these experiments, we aim to gain insights into the CNNs' potential to extract and leverage color-invariant features, thereby enhancing their generalization ability across different color distributions.\par
Overall, this research aims to shed light on the behavior and capabilities of CNNs in the context of color-invariant feature extraction and classification. The findings have implications for understanding the underlying mechanisms of CNNs and can contribute to the development of more robust and adaptable models in various image classification tasks.

\section{Related Work}

Lately, there have been a lot of studies assessing the robustness of deep learning models for the ability to generalize for out-of-distribution testing data. A study by Geihros et al \cite{texturebiasgeirhos} showed that ImageNet-trained CNNs are biased more towards the texture of an object in the image rather than the shape of the object and hypothesized that improving shape bias will improve the performance of CNNs on testing distributions. This hypothesis further investigated in \cite{discriminative} by Islam et al by studying the latent representations to understand at what capacity a CNN learns the shape of the object. They concluded that a CNN learns the shape cues in the first few epochs and that information is mostly stored in the last few layers of a CNN. In \cite{chaitanya} by Chaitanya et al further analyze whether shape bias leads to better robustness and conclude that using different stylized augmentations improve corruption robustness but shape bias is just a byproduct Hosseini et all[4]\cite{shapebiashosseini}\cite{limitationhosseini} conducted several experiments on MNIST and CIFAR10 \cite{cifar} data to assess the shape bias property of CNNs. They created a complimented dataset of MNIST images by taking a negative of original images thereby preserving the shape cues but changing the background and foreground of an image. Their experiments showed that CNNs trained on normal images lack the ability to detect negative images. However, mixing a few negative images with the normal images helped the neural network to learn to detect negative images. In this study, we focus on studying the ability of CNNs to extract intensity-based features for classification by maintaining the intensity of foreground constant across the datasets. A few recent studies on the spatial shift-invariance of CNNs concluded that CNNs are not shift-invariant by design but can learn to be shift-invariant by proper data augmentation and initialization \cite{Zhang}.
\section{Experimental Setup}

\subsection{Procedure}

\begin{figure}[h]
     \centering
     \begin{subfigure}[b]{0.3\textwidth}
         \centering
         \includegraphics[width=\textwidth]{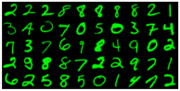}
         \caption{MD1}
         \label{fig: MNIST_green}
     \end{subfigure}
     \hfill
     \begin{subfigure}[b]{0.3\textwidth}
         \centering
         \includegraphics[width=\textwidth]{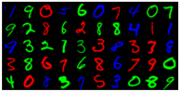}
         \caption{MD2}
         \label{fig:MNIST_rvgvb}
     \end{subfigure}
     \hfill
     \begin{subfigure}[b]{0.3\textwidth}
         \centering
         \includegraphics[width=\textwidth]{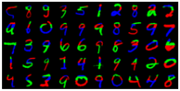}
         \caption{MD3}
         \label{fig: MNIST_rgb}
     \end{subfigure}
        \caption{Modified MNIST datasets}
        \label{fig: modified_mnist}
\end{figure}

As mentioned earlier, the MNIST handwritten data is utilized to generate three different distributions for testing our models. MNIST is a large database of handwritten digits commonly employed for training various image processing systems. The database comprises 60,000 training images and 10,000 testing images. The first dataset, referred to as MD1, is created using exclusively green-colored images, as depicted in Algorithm \ref{alg:green_colored_mnist}. To generate a dataset of single-channeled images (MD2), we randomly select a channel (red, green, or blue) and copy the corresponding pixel values from the MNIST data, as illustrated in Algorithm \ref{alg:random_color_mnist}. Finally, to create a dataset of three-channel images (MD3), we generate a tuple of three numbers to represent the RGB channels, permute them randomly, and copy the pixel values from the MNIST data to their respective channels \ref{alg:colored_mnist_horizontal}. This approach ensures that the intensities of corresponding pixels remain unchanged across all the datasets, thereby preventing the establishment of a direct correlation between color and digit. Figure \ref{fig: modified_mnist} showcases some examples from the aforementioned datasets.

\begin{algorithm}[h!]
    \caption{Generating Green Colored MNIST Images}
    \label{alg:green_colored_mnist}
    \textbf{Input:} MNIST dataset \\
    \textbf{Output:} Green color images
    \begin{algorithmic}[1]
        \For{each image in the MNIST dataset}
            \State Reshape the image to a $32 \times 32 \times 1$ tensor
            \State Create an empty tensor with zeros in the shape $32 \times 32 \times 3$
            \State Set the Green channel values to a fixed green value (e.g., 255)
            \State Save the colored image to disk
        \EndFor
    \end{algorithmic}
\end{algorithm}

\begin{algorithm}[h!]
    \caption{Randomly Coloring MNIST Images to Red, Blue, or Green Only}
    \label{alg:random_color_mnist}
    \textbf{Input:} MNIST dataset \\
    \textbf{Output:} Red, Blue, or Green colored images
    \begin{algorithmic}[1]
        \State \textbf{for} each image \textbf{do}
        \State \quad Reshape the image to $32 \times 32 \times 1$ tensor
        \State \quad Choose a random channel: $0, 1, 2$
        \State \quad Create an empty $32 \times 32 \times 3$ tensor
        \State \quad Set the values of the selected channel to the original pixel values
        \State \quad Set the values of the other channels to 0
        \State \quad Save the colored image to disk
        \State \textbf{end for}
    \end{algorithmic}
\end{algorithm}

\begin{figure}[t!]
     \centering
     \begin{subfigure}[t]{0.3\textwidth}
         \centering
         \includegraphics[width=\textwidth]{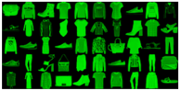}
         \caption{FD1}
         \label{fig: FashionMNIST green}
     \end{subfigure}
     \hfill
     \begin{subfigure}[t]{0.3\textwidth}
         \centering
         \includegraphics[width=\textwidth]{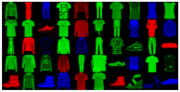}
         \caption{FD2}
         \label{fig: FashionMNIST rvgv}
     \end{subfigure}
     \hfill
     \begin{subfigure}[t]{0.3\textwidth}
         \centering
         \includegraphics[width=\textwidth]{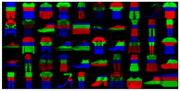}
         \caption{FD3}
         \label{fig: FashionMNIST rgb}
     \end{subfigure}
        \caption{Modified FashionMNIST datasets}
        \label{fig: modified fashionmnist}
\end{figure}

\begin{algorithm}[h!]
    \caption{Generating Colored MNIST Images with Equal Parts of Red, Green, and Blue Horizontally}
    \label{alg:colored_mnist_horizontal}
    \textbf{Input:} MNIST dataset \\
    \textbf{Output:} Images with Equal Parts of Red, Green, and Blue Horizontally
    \begin{algorithmic}[1]
        \State \textbf{for} each image in the MNIST dataset \textbf{do}
        \State \quad Reshape the image to $32 \times 32 \times 1$ tensor
        \State \quad Create a zero-filled $32 \times 32 \times 3$ tensor
        \State \quad Generate a random permutation of [1, 2, 3]
        \State \quad Divide the tensor horizontally into three equal parts
        \State \quad Fill the first third of the tensor with the original pixel values in the channel corresponding to the first element of the permutation
        \State \quad Fill the second third of the tensor with the original pixel values in the channel corresponding to the second element of the permutation
        \State \quad Fill the final third of the tensor with the original pixel values in the channel corresponding to the third element of the permutation
        \State \quad Save the colored image to disk
        \State \textbf{end for}
    \end{algorithmic}
\end{algorithm}
The FashionMNIST dataset is employed to generate distributions comparable to the previously constructed datasets, in order to assess the consistency of our results. FashionMNIST comprises a vast collection of grayscale images depicting various clothing and footwear items. Utilizing the same algorithm described above, we generated three variants of the FashionMNIST data. These variants consist of 'green only images' (referred to as FD1), 'red, green, or blue' images (referred to as FD2), and 'red, green, and blue' images (referred to as FD3). Figure \ref{fig: modified fashionmnist} showcases a sample from each of these datasets.
\subsection{Network Architecture}
\par In this paper, we utilize a standard CNN architecture inspired by VGG16, but with the inclusion of normalization layers, for all the experiments conducted and described. The architecture consists of three convolutional layers with ReLU activations, followed by two dense layers, and a decision layer with softmax activation to classify the images. Similar to the original VGG16, we employ $3x3$ kernels for convolution and $2x2$ kernels for max pooling.\par

\begin{figure}
     \centering
    \includegraphics{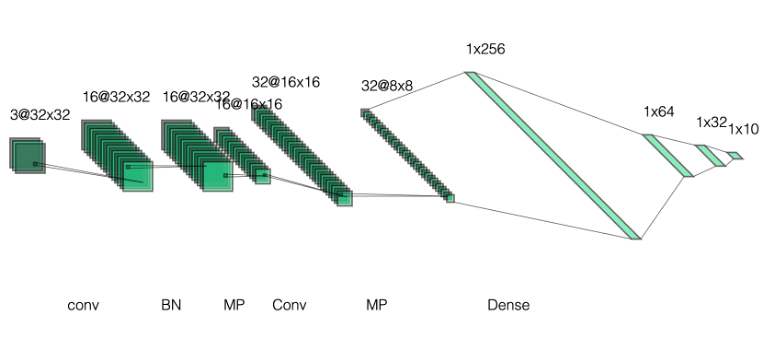}
    \caption{Network Architecture}
    \label{fig:net arch}
\end{figure}

This modified VGG16 model contains a total of $1500000$ parameters. It offers a balance between ease of training and testing on the modified datasets while providing sufficient capacity for accurate classification. The model architecture is shown in Figure \ref{fig:net arch}

\par Three identical models were created with the same initialization to ensure that the role of the initialization is nulled in the experiments. All the models were trained for 50 epochs and the weights with the best validation accuracy are stored. The first model, the ‘MM1’, is trained on 'MD1', containing only green-colored images. The second model 'MM2', is trained on the second dataset 'MD2',  containing only images having one color channel from red, green, or blue. The third model, the ‘MM3’ is trained on the third dataset 'MD3' with images having all three-color channels. Similarly 'FM1', 'FM2', and 'FM3' are the models trained on 'FD1', 'FD2', and 'FD3' respectively. All models were tested on all the datasets to measure their ability to make decisions on shape cues.

To eliminate the influence of initialization, three identical models were created and initialized with the same weights for the experiments. Each model was trained for 50 epochs, and the weights corresponding to the best validation accuracy were saved. The first model, named 'MM1', was trained on the 'MD1' dataset consisting of only green-colored images. The second model, 'MM2', was trained on the 'MD2' dataset, which includes images with a single color channel randomly chosen from red, green, or blue. The third model, 'MM3', was trained on the 'MD3' dataset, which contains images with all three color channels. Similarly, three models named 'FM1', 'FM2', and 'FM3' were trained on the 'FD1', 'FD2', and 'FD3' datasets, respectively, derived from the FashionMNIST data. All models were tested on all their respective datasets to evaluate their ability to make decisions based on shape cues.

\section{Results and Discussion}

\subsection{Effect on other distributions}

In this section, we investigate whether CNNs trained exclusively on green-colored images can accurately classify the same images when presented in different colors. The 'green model' is trained on the MNIST data for 50 epochs, and the weights corresponding to the best validation loss are saved. Similarly, the other two models are initialized and trained on their respective datasets. Subsequently, all models are tested on all versions of the test datasets. The results are shown in Table \ref{tab:table1}.

\begin{table}
	\caption{Model accuracy on Modified MNIST datasets}
	\centering
	\begin{tabular}{c|c|c|c}
	\hline
		   & Accuracy on MD1     & Accuracy on MD2     & Accuracy on MD3 \\
		\midrule
		MM1   & $0.99$        & $0.40$   &  $0.21$ \\
		MM2   & $0.98$        & $0.99$    & $0.97$ \\
		MM3     & $0.98$       & $0.99$   & $0.99$ \\
		\bottomrule
	\end{tabular}
	\label{tab:table1}
\end{table}
These experiments showed that although the model trained on the MD1 dataset, achieved $99\%$ accuracy on the MD1 test data containing only green images, its accuracy dropped significantly when tested on the other variants of the same data. The model's accuracy on the MD2 dataset containing red, green, or blue images is $40\%$ because $33.33\%$ of test images are similar to MD1. This reinforces our assumption that the model did not learn the color invariant intensity-based features while training on the green only images. The model's accuracy statistic further decreased when tested on the third dataset containing images from all three colors, achieving only a paltry $21\%$. The second model 'MM2' trained on the MD2 dataset containing red, green or blue images achieved similar accuracy of $99\%$ on the test dataset of its variant but also achieved $99\%$ on the MD1 dataset as well. However, its accuracy dropped by $2\%$ achieving $97\%$ when tested on the images containing all three colors. This could be because although the model is presented with images from all the color channels while training, they did not occur together at the same time.
The experimental findings demonstrated that the model trained on the MD1 dataset exhibited exceptional accuracy of $99\%$ when classifying green-only images. However, its performance significantly deteriorated when confronted with datasets containing different color distributions. Specifically, the model's accuracy on the MD2 dataset, consisting of red, green, or blue images, dropped to $40\%$, and further plummeted to a mere $21\%$ on the MD3 dataset, which encompassed images with all three color channels.
Likewise, the MM2 model, trained on the MD2 dataset, showcased impressive accuracy on both the MD2 and MD1 datasets. Nevertheless, it experienced a slight decline, achieving an accuracy of $97\%$, when tested on images containing all three color channels.
These outcomes underscore the inherent challenge of training models to learn color-invariant features. It is evident that models trained exclusively on specific color distributions struggle to generalize effectively to different color variations. This highlights the need for robust techniques to enhance color invariance and bolster the model's performance across diverse color combinations, thereby augmenting its practical utility.

\par The models were trained and tested on the FashionMNIST variants of the data to see if the  results were consistent with the ModiefiedMNIST variants. The results are shown in the Table \ref{tab:table1}.\par

\begin{table}
	\caption{Model accuracy on Modified FashionMNIST datasets}
	\centering
	\begin{tabular}{c|c|c|c}
	\hline
		   & Accuracy on FD1     & Accuracy on FD2     & Accuracy on FD3 \\
		\midrule
		FM1   & $0.91$        & $0.49$   &  $0.25$ \\
		FM2   & $0.91$        & $0.90$    & $0.85$ \\
		FM3     & $0.91$       & $0.90$   & $0.91$ \\

   \bottomrule
	\end{tabular}
	\label{tab:table1}
\end{table}

\par The model results were comparable to the results on the MNIST data, with the 'FM1' model failing to classify images containing other color channels. This result supports the assumption that color and shape are interrelated in the RGB color model and that a filter that extracts a specific shape may not be helpful for extracting the same shape in a differently colored image.\par

\subsection{Effect of Normalization}

\par Batch normalization is a widely used technique in a model building that aims to address the issue of internal covariate shift. This phenomenon refers to the significant change in layer activations that can occur due to variations in the parameter values of a previous layer \cite{batch}. By normalizing the inputs within each mini-batch during training, batch normalization helps stabilize and improve the learning process.

While batch normalization has been the go-to choice for normalizing neural network activations, other normalization techniques have emerged in recent years. Two notable alternatives are Instance normalization \cite{instance} and Layer normalization \cite{layer}. Instance normalization, often employed in generative models, normalizes the activations of each instance individually, making it well-suited for tasks involving style transfer or image generation. On the other hand, Layer normalization normalizes the activations within each layer across all instances, offering a different approach to address the covariate shift problem.

In this section, we aim to investigate how different normalization techniques impact the performance of our model when dealing with various color distributions. To accomplish this, we replaced the batch normalization layers used in our previous experiments with Instance normalization and Layer normalization. By conducting the same set of experiments with these alternative normalization techniques, we can gain insights into their effectiveness and compare them against batch normalization.

By examining the results obtained from these experiments, we can evaluate the performance of the models and determine which normalization technique proves the most effective in handling the challenges posed by different color distributions. The results of these experiments are shown in  Table \ref{tab:table3}.\par

\begin{table}
	\caption{Accuracy on Modified MNIST datasets using different normalization layers}
	\centering
	\begin{tabular}{c|c|c|c}
	\hline
		   & Accuracy on MD1     & Accuracy on MD2     & Accuracy on MD3 \\
		\midrule
		BN model   & $0.99$        & $0.40$   &  $0.21$ \\
		LN model   & $0.99$        & $0.56$    & $0.47$ \\
		IN model     & $0.99$       & $0.50$   & $0.43$ \\
		\bottomrule
	\end{tabular}
	\label{tab:table3}
\end{table}

\par When layer normalization was used instead of batch normalization the performance of the model increased by $16\%$ on the second dataset and almost $100\%$ on the third dataset. Using instance normalization instead of batch normalization improved the accuracy significantly on 'MD2' data and 'MD3' data. With the same choices in the experiments on FashionMNIST, the results were slightly different, as shown in Table \ref{tab:table4}.\par
\par Upon replacing batch normalization with layer normalization, remarkable improvements were observed in the model's performance. Specifically, there was an increase of $16\%$ in accuracy when evaluating the model on the second dataset. Additionally, when testing the model on the third dataset, an outstanding improvement of almost $26\%$ in accuracy was achieved. These findings indicate that layer normalization proved to be a superior normalization technique compared to batch normalization for handling the challenges posed by different color distributions. \par
\par Contrary to our initial expectations, when instance normalization was utilized instead of batch normalization, the observed improvements in accuracy were not as substantial as with layer normalization. While there were notable enhancements in accuracy for the datasets containing red, green, and blue images, the overall performance was not as impressive as with layer normalization. Furthermore, when the model was tested on datasets combining all three color channels, the improvements achieved with instance normalization were not as significant.
Hence, based on our experiments and results, we recommend utilizing layer normalization as the preferred normalization technique when dealing with models that need to handle diverse color distributions.\par
\begin{table}
	\caption{Accuracy on Modified FashionMNIST datasets using different normalization layers}
	\centering
	\begin{tabular}{c|c|c|c}
	\hline
		   & Accuracy on FD1     & Accuracy on FD2     & Accuracy on FD3 \\
		\midrule
		BN model   & $0.91$        & $0.49$   &  $0.25$ \\
		LN model   & $0.92$        & $0.61$    & $0.62$ \\
		IN model     & $0.92$       & $0.57$   & $0.49$ \\
		\bottomrule
	\end{tabular}
	\label{tab:table4}
\end{table}

\subsection{Effect of Architecture}
\par In order to investigate the impact of architecture on the extraction of color-invariant features for classification, we conducted tests using various standard architectures. The models were evaluated on both the MNIST and FashionMNIST datasets, encompassing their respective color variants. The results obtained from these experiments are presented below in table \ref{tab:table4} and table \ref{tab:table5}.

It is worth noting that while some models exhibited a marginal improvement in accuracy, the observed enhancements were not substantial enough to be considered practically significant. These findings suggest that the choice of architecture alone may not play a major role in effectively extracting color-invariant features for classification tasks.

These results emphasize the need to explore additional strategies and techniques beyond architecture selection to address the challenge of color invariance in image classification. It is crucial to develop more sophisticated approaches that can effectively capture and utilize color-independent features, enabling models to achieve robust performance across diverse color distributions. The results are shown in the tables 

\begin{table}
	\caption{Accuracy on Modified MNIST datasets using standard architectures}
	\centering
	\begin{tabular}{c|c|c|c}
	\hline
		   & Accuracy on MD1     & Accuracy on MD2     & Accuracy on MD3 \\
		\midrule
		vgg16   & $0.99$        & $0.39$   &  $0.32$ \\
		mobilenetv2    & $0.98$        & $0.49$    & $0.43$ \\
		resnet50     & $0.99$       & $0.43$   & $0.41$ \\
		\bottomrule
	\end{tabular}
	\label{tab:table4}
\end{table}
\begin{table}
	\caption{Accuracy on Modified FashionMNIST datasets using standard architectures}
	\centering
	\begin{tabular}{c|c|c|c}
	\hline
		   & Accuracy on FD1     & Accuracy on FD2     & Accuracy on FD3 \\
		\midrule
		vgg16   & $0.92$        & $0.37$   &  $0.32$ \\
		mobilenetv2    & $0.91$        & $0.40$    & $0.23$ \\
		resnet50     & $0.92$       & $0.43$   & $0.41$ \\
		\bottomrule
	\end{tabular}
	\label{tab:table5}
\end{table}

\subsection{Confidence in Prediction}

\par To examine the relationship between class labels and correct predictions, we conducted an analysis of the outputs generated by the green model on different distributions of the dataset. The findings of this analysis are presented in \ref{tab:confidence table}.

From \ref{tab:confidence table}, it becomes evident that the model exhibits higher confidence levels in its incorrect predictions when tested on datasets that deviate from the training distribution. Interestingly, the confidence associated with correct predictions is slightly lower than that of the incorrect predictions on the 'MD1' dataset. This counter-intuitive observation suggests that the inclusion of information from other color channels has led the model to confidently make erroneous predictions.

These results shed light on the complex interplay between color distributions and the model's decision-making process. The higher confidence observed in incorrect predictions on out-of-distribution datasets underscores the model's vulnerability to variations in color. This highlights the importance of developing strategies to enhance the model's robustness to color variations, allowing for more accurate and reliable predictions.

Further investigation and research are needed to understand the underlying factors contributing to these observations and to develop techniques that effectively address the challenges posed by color variations in image classification tasks. By gaining a deeper understanding of these dynamics, we can advance the field and devise more robust models capable of achieving improved performance across diverse color distributions..\par

\begin{table}[]
  \centering
  \caption{Prediction probabilities}
  \begin{tabular}{c|cc|cc|cc}
    \toprule
    \textbf{class\_labels} & \textbf{MD1 correct} & \textbf{MD1 incorrect} & \textbf{MD2 correct} & \textbf{MD2 incorrect} & \textbf{MD3 correct} & \textbf{MD3 incorrect} \\
    \midrule
    0 & 0.999 & 0.860 & 0.998 & 0.962 & 0.836 & 0.929 \\
    1 & 0.999 & 0.893 & 0.974 & 0.935 & 0.829 & 0.940 \\
    2 & 0.998 & 0.807 & 0.969 & 0.942 & 0.981 & 0.947 \\
    3 & 1.000 & 0.890 & 0.977 & 0.942 & 0.967 & 0.925 \\
    4 & 0.999 & 0.872 & 0.996 & 0.938 & 0.977 & 0.943 \\
    5 & 0.999 & 0.957 & 0.986 & 0.949 & 0.962 & 0.933 \\
    6 & 0.998 & 0.831 & 0.996 & 0.945 & 0.874 & 0.941 \\
    7 & 0.999 & 0.669 & 0.986 & 0.946 & 0.959 & 0.948 \\
    8 & 0.996 & 0.841 & 0.996 & 0.941 & 0.900 & 0.952 \\
    9 & 0.994 & 0.904 & 0.984 & 0.930 & 0.919 & 0.935 \\
    \bottomrule
  \end{tabular}
  \label{tab:confidence table}
\end{table}

\subsection{Correlation between class label and number of right predictions}

\par The analysis revealed a notable correlation between the class labels and the number of correct predictions across the datasets. Despite ensuring no direct correlation between color and class label, the results indicated a strong association. Class label '4' consistently achieved a high number of correct predictions, while class label '0' had the lowest accuracy. These findings suggest the presence of underlying factors influencing classification accuracy beyond color. Further investigation is necessary to uncover these factors and improve overall classification performance.\par

\begin{figure}
     \centering
    \includegraphics[scale = 0.5]{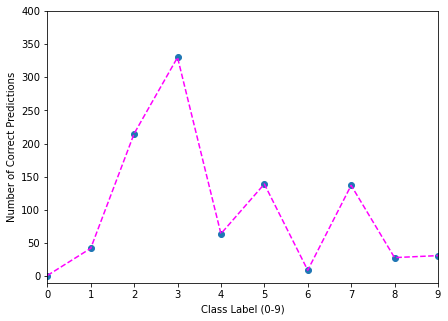}
    \caption{graph showing class label and number of right predictions}
    \label{fig:confidence_in_prediction}
\end{figure}
\section{Improving Shape Bias}

\par In order to increase the emphasis on intensity-based features, we implemented a strategy to introduce a grayscale version of the input data. This was accomplished by calculating a weighted average value across the color channels and concatenating it with the original input. Additionally, we designed a specialized dropout layer that selectively assigns zero weights to the input image based on a random variable, as outlined in \ref{alg: custom dropout}

\begin{algorithm}
\caption{Custom Dropout}
\begin{algorithmic}[1]
\Require \textit{inputs} (input tensor), \textit{prob} (dropout probability)
\Ensure \textit{output} (output tensor)

\State \textbf{Initialization:}
\State $rand\_prob \gets \textnormal{random.uniform}([])$ 
\State $true\_mask \gets \textnormal{Concatenate}(axis=3)([\textnormal{zeros}([1,32,32,3]), \textnormal{ones}([1,32,32,1])])$ 

\Procedure{CustomDropout}{$inputs$, $prob$}
    \If{$rand\_prob < prob$}
        \State \textbf{return} $inputs \times true\_mask$ \Comment{Element-wise multiplication}
    \Else
        \State \textbf{return} $inputs$
    \EndIf
\EndProcedure

\State \textbf{Usage:}
\State \textbf{Replace} $inputs$ \textbf{with the actual input tensor}, and $prob$ \textbf{with the desired dropout probability}
\State $output \gets \textnormal{CustomDropout}(inputs, prob)$

\end{algorithmic}
\label{alg:custom_dropout}
\end{algorithm}

Through a series of experiments, we evaluated the performance of this modified architecture on both the MNIST and FashionMNIST datasets. The results obtained were highly promising, demonstrating a significant improvement in classification accuracy compared to the baseline models. The specific accuracies achieved are provided in the accompanying table, which highlights the notable gains attained through the incorporation of grayscale information and the utilization of the customized dropout layer.

These findings have important implications, as they underscore the effectiveness of integrating intensity-based features into the classification process. Furthermore, they emphasize the value of leveraging tailored dropout mechanisms to enhance the discriminative capabilities of the network.
\section{Conclusion}
\par In conclusion, our experiments focused on investigating the impact of color variations on the performance of convolutional neural networks (CNNs) for image classification tasks. We utilized modified versions of the MNIST and FashionMNIST datasets, which allowed us to examine the models' ability to extract shape cues while being invariant to color changes.

Our findings indicate that CNNs trained on specific color distributions, such as green-only or red, green, or blue images, exhibit a strong reliance on the training color and struggle to generalize to other color distributions. This suggests that the models fail to capture color-invariant intensity-based features during training.

We explored the effects of different normalization techniques on the models' performance, such as batch normalization, instance normalization, and layer normalization. The experiments revealed that layer normalization outperformed instance normalization, demonstrating its effectiveness in improving the models' accuracy across different color distributions.

Furthermore, we introduced an enhanced model architecture that incorporated grayscale computation and a custom dropout layer. This modification aimed to increase the model's bias toward intensity-based features and enhance its ability to extract color-invariant shape cues. The results demonstrated the efficacy of this approach, as the improved model consistently achieved higher accuracy on all datasets compared to the baseline models.

Overall, our study highlights the importance of considering color variations and the need for models that can effectively extract shape cues while being invariant to color changes. Our findings contribute to a deeper understanding of the challenges posed by color variations in image classification tasks and provide valuable insights for developing more robust and color-invariant CNN models in the future.\par

\bibliographystyle{unsrt}  
\bibliography{references}

\end{document}